%% file: Nemlap.tex
\documentstyle[psfig]{nemlap}

\begin{document}

\title{Storage of Natural Language Sentences in a Hopfield Network}
\author{Nigel Collier\footnote{Published in Proceedings of the International Conference on New Methods in Natural Language Processing (NeMLaP-2), Bilkent University, Ankara, Turkey, September 16--18 1996.}\footnote{While any mistakes in content are entirely my own, I would like to thank my supervisor Professor Tsujii for his many helpful and critical observations. I also gratefully acknowledge the kind permission of Asahi Newspapers for the use of their editorial corpus. Funding was provided by the Economics and Social Research Council in the UK award no. R00429434065. Finally I would like to thank the reviewers for their insightful comments.}}
\institute{Department of Language Engineering\\
UMIST\\
P.O.Box 88, Manchester M60 1QD, UK\\
E-mail: nigelc@ccl.umist.ac.uk}

\maketitle

\begin{abstract}
This paper look at how the Hopfield neural network can be used to store and recall patterns constructed from natural language sentences. As a pattern recognition and storage tool, the Hopfield neural network has received much attention. This attention however has been mainly in the field of statistical physics due to the model's simple abstraction of spin glass systems.
 
A discussion is made of the differences, shown as bias and correlation, between natural language sentence patterns and the randomly generated ones used in previous experiments. Results are given for numerical simulations which show the auto-associative competence of the network when trained with natural language patterns.
\end{abstract}

\section{Introduction}

As a pattern recognition and storage tool, the Hopfield network \cite{hopfield:82,hopfield:84} has received much attention. In particular the discrete model has been widely investigated in the field of statistical physics (e.g. \cite{amit:87,amit:89} \cite{bruce:87} \cite{forrest:88} \cite{gardner:86,gardner:87} \cite{nishimori:93} \cite{ozeki:94}) because of its simple abstraction of a spin glass system.

This paper looks at how the Hopfield memory can be used to store and recall patterns which represent natural language sentences. It will be shown that these patterns behave differently to those randomly generated ones used in previous experiments.

The feature of natural language patterns which we describe in this paper is that the networks they form have very low activation levels. This introduces complications from a computational point of view in the form of non-zero mean noise, which unchecked would make recall impossible. From an AI perspective we note that low activation is an intrinsic feature of the parts of the brain dealing with concept association.

Firstly though, why use the Hopfield network in NLP? The goal of our research is to adapt associative neural networks for use in context related NLP tasks such as word sense disambiguation and lexical transfer in machine translation. It is generally agreed that contextual knowledge plays an important role in the processing of language by people, but the complexity of word relations which together represent \lq context' has defied analysis and prohibited progress towards context driven processing in NLP. Our intention is to use the neural network as a sophisticated post-processing device in which contextual knowledge can be utilised through a multiplicity of word cooccurrence relations in a fully connected network. This contrasts with traditional statistical NLP methods such as those in \cite{brown:93} and \cite{gale:93b} which tend to model context through local surface word cooccurrence n-grams.

One of the weaknesses of such statistical methods is in their localist treatment of context where only cooccurrences in a narrow window, usually covering a single sentence, contribute to the modelling of context. This ignores the fact that much contextual knowledge may be outside the sentence in which a word occurrs. We refer to such external contextual relations as \lq global context' and our network approach seeks to capture this through indirect word association relations and to process it efficiently.

Our basic criteria for choosing a connectionist approach are (a) automatic knowledge acquisition, (b) avoidance of combinatorial explosion, and (c) knowledge transparency.

We view the need for machine learning of language from examples and a self-organising memory as crucial to large scale NLP. In this respect we agree with other so called \lq bottom-up' paradigms such as statistical NLP and example-based NLP which are based on automatic knowledge acquisition. Unlike statistical NLP however, we see that the mathematical tools for measuring surface word associations in large corpora will yield only rough approximations for most of the interesting word relations and look for a better way to process the statistical knowledge with all its inaccuracies.

Combinatorial explosion is a problem for large scale NLP with most paradigms and makes scaling up of systems difficult. We hope to be able to avoid this by using a model in which the number of word relations, as measured by simple coooccurrences in sentences, do not effect efficiency in terms of processing times.

Transparency of the knowledge base is our final basic criteria and means that the representation of linguistic knowledge should be analysable and easily interpreted by non-experts in connectionism. This should aid verification of results.

Neural networks are thought of as being \lq black boxes' in which knowledge is so heavily encoded that it defies inspection. We will be using a localist storage method in which semantic transparency of language is preserved. Additionally, the Hopfield network has dynamics which are mathematically tractable and understood within the limits imposed by statistical physics. We can therefore apply both a linguistic as well as a mathematical explanation to our results. This contrasts with previous work in connectionist NLP such as Ide and Veronis \cite{ide:90} where the networks used are structured and incomplete and have no well understood theoretical framework.

In addition to our basic criteria we look particularly to neural networks to provide generalisation. This is a crucial capability in any paradigm which processes language because we cannot expect to train our models on the complete set of examples which we will encounter. The generalization capability manifests itself in the network being able to learn relationships which were not linear in the training set. We hope to exploit this function and develop a practical connectionist alternative to other bottom-up NLP paradigms.

As a first step in an ongoing series of experiments we intend to explore the basic functionality of the Hopfield network for use in storing natural language sentences. This is important because it establishes the basic properties of the network and provides a foundation for future work in this area. In the future we want to develop the model to allow multi-word sense disambiguation in sentences using large-scale contextual knowledge derived from corpus statistics.

\section{Sentence Patterns}

Previous work in statistical physics has looked at the storage of randomly generated bit vectors in which the bits all have equal probabilities of being 1 or 0. For many real world tasks such as NLP or vision processing, we would like to store non-random patterns where the bits do not occur with equal likelihood.  It is therefore natural to explore the properties of the model for storing non-random patterns in our domain of interest.

The training vectors we use can be regarded as a set of $n$ patterns $\xi^{(\mu)}$ for ($\mu = 1,..,n$) representing the $n$ sentences we wish to store. Each pattern consists of $N$ bits with each bit taking a value of 0 or 1, where $\xi^{\mu}_{i}=1$ ($\xi^{\mu}_{i}=0$) represents the presence (absence) of a word at index $i$ in a lexicon in pattern (and sentence) $\mu$. In our localist representation each unit in the Hopfield network derived from these training patterns similarly corresponds to one word in the lexicon.

The representation allows us to model linguistic properties of language which are dependent on frequency and cooccurrence of words. With enough training data we can capture useful information about word contexts.

Several linguistic factors make natural language patterns different to those generated randomly. These are:

\begin{enumerate}
\item Bias. Words in the lexicon do not occur with the same distribution. Technical vocabulary and proper names especially tend to have a very low frequency, even when the training corpus is very large.
\item Internal Correlation: Syntactic and semantic factors mean that the probability distributions of two words appearing in the same sentence are not independent.
\item External Correlation: Pragmatic factors mean that words and phrases which appear before others in a continuous and coherent piece of text influence the probability distributions of those which come later.
\end{enumerate}

It is beyond the scope of this work to calculate a macro-statistic for correlation between patterns as described above. We can however think of other measures which reflect part of this information. These are outlined below.

Bias is a measure of how likely a bit in the training set $\xi^{(\mu)}$ is to be 1. In our model this is  

\begin{equation}
Pr(\xi^{(\mu)}_{i} = 1) = \frac{\sum^{n}_{\mu}\sum^{N}_{i}\xi^{\mu}_{i}}{n N}
\label{Bias}
\end{equation} 

This statistic serves only as a gross summary and does not show correlation features of patterns which were discussed above. Nevertheless it is simple to calculate and shows us something of the nature of the patterns we are storing. Clearly for a network storing unbiased patterns $Pr(\xi^{(\mu)}_{i}=1)$ will be 0.5.

Pattern recognition studies which use randomly generated patterns have for practical reasons assumed that bias is the same for all bits of all patterns in all contexts. For natural language patterns we should not make such an assumption because the frequency of bits is directly linked to the distribution of words in a corpus. 

Previous work (e.g. \cite{amit:89} and \cite{watkin:91}) has shown that connectivity is also an important factor in determining the network's behaviour.
We will define mean connectivity informally as the mean number of different words which any single word cooccurs with in a sentence. We will define this more formally in the next section.

\section{The Model}

The discrete Hopfield model which we explore as the basis for our work is a fully connected network of {\it N} units, where the synaptic connection strengths are held in a \lq weight' matrix {\it T} with $T_{ij}$ representing the weight on the symmetrical arc between units $i$ and $j$. The output from a unit {\it i} is $V_{i}$ and comes from internal and external sources with the internal inputs

\begin{equation}
H_{i} = \sum_{j, j \neq i}^{N} T_{ij} V_{j} - U_{i}
\label{InternalInputs}
\end{equation}

where $U_{i}$ is a threshold. The external input $I_{i}$ is calculated and set at the start of processing. Note that self interaction between a unit and itself is prohibited.
 
In the version of the network which we use, stored patterns are recalled using a {\it recall prescription} 

\begin{equation}
V_{i} = \left\{ \begin{array}
        {r@{\quad if \quad}l}
        0 & N_{i} < U_{i}\\
        1 & N_{i} \geq U_{i}\\
        \end{array} \right.
\label{Output}
\end{equation}

where

\begin{equation}
N_{i} = \sum_{j, j \neq i}^{N} T_{ij} V_{j} + I_{i}
\label{InputSum}
\end{equation}

The operation of individual units in the network is quite simple as we can see from Eqn. (\ref{Output}), where a weighted sum of inputs from all other units determines whether the unit outputs a 1 or a 0. This disguises the fact that the collective behaviour of a system of such fully connected units is quite complex.

Training the network occurs by bringing into correspondence the patterns we wish to store, $\xi^{(\mu)}$ for ($\mu = 1,..,n$), and stable states in the network's dynamics, $V^{(\mu)}$, called {\it nominated states}. At the same time we want to avoid creating spurious stable states which do not correspond to any of the training patterns. 

Storage is effected through the weight matrix {\it T} and the threshold vector {\it U}. Although there are more effective storage prescriptions (e.g. see Tarassenko {\it et al} \cite{tarassenko:90}), we have chosen to use the Hebb rule 

\begin{equation}
T_{ij} = \frac{1}{N} \sum_{\mu}^{n} \xi_{i}^{\mu} \xi_{j}^{\mu}
\label{StoragePrescription}
\end{equation}

and to set all the elements in the threshold vector, {\it U}, to a constant which is calculated at the start of processing. Using the Hebb rule makes our results compatible with earlier work in statistical physics and is also computationally convenient when computing the weight matrix. 

For storage to be guaranteed to take place, {\it T} must be both symmetrical and have a zero diagonal, so we introduce the additional rule

\begin{equation}
T_{ii} = 0 
\end{equation}

The Hebb rule (\ref{StoragePrescription}) for finding the weights between two units (words) in the network intuitively corresponds to the frequency of cooccurrence of the words in the training sentences, ignoring multiple cooccurrence in the same sentence. This simple relation between the training data and the representation ensures semantic transparency. We can also see that learning of sentences in the approach we have outlined here is both automatic and self-organizing, in that we do not decide a priori which word relations are to remain and which are to be ignored.

Unfortunately, mean field analysis by Amit \cite{amit:89} has predicted that using the storage equation (\ref{StoragePrescription}) with biased patterns will lead to noise overwhelming signal and the destabilising of nominated states. Rather than use a non-localist storage prescription such as the projection matrix of Kohonen \cite{kohonen:73} or Personnaz {\it et al} \cite{personnaz:85} we intend to compensate  for local noise by implementing a global inhibitor. This is inspired by comments in Buhmann and Schulten \cite{buhmann:87} as well as Amit \cite{amit:89} and allows us to stabilise states corresponding to nominated patterns by compensating for the noise. In this way we have replaced elements $T_{ij} = 0$ where $i \neq j$ with $T_{ij} = -10/N$. The constant 10 corresponds approximately to the number of content words (the number of 1s) in a training pattern vector.

We can now formally define mean connectivity, $\overline{c}(T)$ as

\begin{equation}
\overline{c}(T) = N^{-1}\sum^{N}_{i} \sum^{N}_{j} g(T_{ij})
\label{Connectivity}
\end{equation}

for 

\begin{equation}
g(x) = \left\{ \begin{array}
        {r@{\quad if \quad}l}
        1 & x \geq 1\\
        0 & otherwise\\
        \end{array} \right.
\end{equation}

Finally we define matrix sparsity which shows the fraction of the weight matrix $T$ which is non-zero

\begin{equation}
s(T) = N^{-1}\overline{c}(T)
\end{equation}

In order to recall the stored patterns the initial output vector is set to a training pattern from $\xi^{(\mu)}$. The network is then updated stochastically and randomly until it settles into a stable state as shown by the convergence of the energy function

\begin{equation}
E(\{V\}) = -\frac{1}{2} \sum_{i}^{N} \sum_{j}^{N} T_{ij} V_{i} V_{j} + \sum_{i}^{N} U_{i} V_{i} - \sum_{i}^{N} I_{i} V_{i}
\label{Lyaponov}
\end{equation}

$E(\{V\})$ has been proven by Hopfield \cite{hopfield:82} to be a strictly decreasing function of processing time and converges when the network has settled into a stable state. 

\section{Limitations}

Numerical studies by Hopfield \cite{hopfield:82} showed that the effective storage capacity of the network was linked to a {\it storage ratio}

\begin{equation}
\alpha = n/N
\label{StorageRatio}
\end{equation}

where {\it n} is the number of patterns and {\it N} is the number of bits in each pattern. Initial estimates for reliable storage showed that a value of $0.1 \leq \alpha \leq 0.2$ was most effective.

Analytical techniques used by Amit {\it et al} \cite{amit:87b} have shown the existence of a critical value of $\alpha$ called $\alpha_{c}$ where auto-associative recall degrades discontinuously when $\alpha$ exceeds $\alpha_{c}$. Further studies by Grensing {\it et al} showed that if we accept a small amount of error, say 0.005\% then $\alpha_{c} \approx 0.15$.

The reason why patterns can be stored and recalled in the Hopfield network is because the patterns are made to correspond to stable states, called {\it minima}  in the energy landscape formed by the set of all possible output states of the network. When $\alpha \leq \alpha_{c}$ each stored pattern corresponds to a single stable state. As the critical value is exceeded multiple correspondences between patterns and stable states occur and spurious minima appear. Recall then rapidly becomes impossible.

For our purposes we view the critical value as a serious limitation to the development of the Hopfield network for practical NLP. This is obvious when we consider that we are limited in the number of patterns we can store to $n \leq N\alpha_{c}$. 

The effect of bias on the critical value through correlated patterns such as those we propose to store is not clear and the literature apparently points in conflicting directions. Amit \cite{amit:87b} for example found that in the large $N$ limit recall degraded discontinuously at $\alpha_{c}$ with random patterns. Researchers who have looked at biased patterns such as Grensing {\it et al} \cite{grensing:87} have found that recall degraded continuously when $\alpha$ exceeded $\alpha_{c}$ up to a second storage ratio value $\alpha_{0}$. Interestingly Amit {\it et al} \cite{amit:87} also found that bias induced a shift in the critical value from 0.14 to 0.18. 

We aim in this paper to show through numerical simulations how bias in natural language sentences effects the critical value and the dynamical properties of the network. In particular we want to see (a) whether non-random biased patterns are stored as effectively as random unbiased ones, and (b) if storage takes place then do we observe a discontinuity and a critical value at $\alpha_{c}=0.14$. We also want to see if we can find some causal relation between bias and the critical storage value.

\section{The Training Corpus}

The training vectors, $\xi^{(\mu)}$, are derived from the Asahi corpus of newspaper editorials. A full specification for this corpus is given by Collier {\it et al} in \cite{collier:95d}, parts of which have been repeated here for completeness. We use this parallel corpus because it is convenient for our next stage of work which will look at lexical transfer, a sort of word sense disambiguation, from English to Japanese. This need not concern us here except in so far as the characteristics of the sentences in the corpus effect the storage results. We therefore present a short outline of the features of the corpus.

The corpus is in English and Japanese and has been aligned at the sentence  level. In our experiments we only use the English sentences. Moreover, as a pre-processing stage we remove all of the function words because they do not contribute significantly to the context of the sentence. Of the 330,000 English words in the corpus, approximately 39.7\% are function words.

The mean length of the sentences which are left is approximately 10 content words and we have calculated that the mean number of cooccurrences between word pairs is 116 . This means that the training patterns are very sparse and so is the resulting weight matrix {\it T}.  

Another influence on the network performance which comes from the training data is the range of values which the storage ratio, $\alpha$, takes. This can be calculated from the lexicon closure curve for the corpus. The lexicon growth curve has been found to closely match the curve $F(n) = 110n^{\frac{1}{2}}$.

If we take the number of sentences as $n$ and the lexicon size as $F(n)$ then we find that as $n$ approaches 12000, $\alpha \approx n/F(n) \rightarrow 1.0$, which is clearly above the values given for $\alpha_{c}$ and indicates that storage of such sentences is impossible. For subcorpora extracted from these 12000 sentences we have generally found that $0.14 \leq \alpha \leq 1.0$.

\section{Results}

In order to test the effectiveness of storage for a particular pattern $\xi^{\mu}$ using numerical simulations we can measure the fractional Hamming distance between an actual stable state, $\tilde{V}^{\mu}$, and the nominated stable state, $V^{\mu}$. This is defined as

\begin{equation}
D_{\mu} = \frac{1}{N} \sum^{N}_{i} | \tilde{V}^{\mu} - V^{\mu} |
\label{HammingDistance}
\end{equation}

Clearly the storage prescription will be effective for a system with $V^{(\mu)}_{i} \in \{0,1\}$ and large {\it N} according to how $D_{\mu}$ is distributed about D=0 over a large number of trials. The following measure of recall effectiveness is derived from Bruce {\it et al} \cite{bruce:87} 

\begin{equation}
\overline{F}_{B} \equiv \sum D_{\mu} \equiv \overline{D}
\label{FB}
\end{equation}

and shows the mean fraction of the nominated images $V^{(\mu)}$ which coincide with their corresponding stable images in $\tilde{V}^{(\mu)}$. i.e. the fraction of bits which are recalled without error over a number of trials. The storage prescription is effective to the extent that $\overline{F}_{B}$ is less than 0.5 -- the value at which there is no coherent overlap between $V^{(\mu)}$ and $\tilde{V}^{(\mu)}$.

To test the effectiveness of storage auto-association tests were done for five test matrices T1 to T5 with specifications given in Table \ref{Tab1}. Auto-association involves presenting the network with a noisy version of a stored pattern and measuring the error in recall according to Eqn. (\ref{FB}). The matrices T1 to T5 represent a range of sizes of subcorpora taken from the Asahi corpus and are expected to show the effects of scale on the Hopfield network.

We may note that since the result of each presentation of a pattern to the model is non-deterministic and we conducted a large number of independent trials we can consider the tests as Monte Carlo simulations. 

We see from the scores for sparsity and connectivity in Table \ref{Tab1} that only a small fraction of the weight matrix is non-zero. This confirms that natural language patterns and the matrices they form are in the class of {\it low activation} networks. Moreover, $\alpha$ exceeds the expected levels for $\alpha_{c} \approx 0.14$ in all matrices.

\begin{table}[h]
\begin{center}
\begin{tabular}{|l | l l l l l|}
\hline
Matrix         & T1    & T2    & T3     & T4     & T5\\
\hline\hline
$N$            & 260   & 673   & 921    & 1357   & 4131\\
$n$            & 41    & 121   & 167    & 273    & 1412\\
$\alpha$       & 0.16  & 0.18  & 0.18   & 0.20   & 0.34\\
$s(T)$         & 0.062 & 0.028 & 0.027 & 0.021 & 0.011\\
$\overline{c}(T)$ & 15.95 & 18.84 & 24.87 & 28.90 & 44.50\\
$Prob(\xi^{(\mu)}_{i}=1)$  & 0.037 & 0.014 & 0.011 & 0.008 & 0.003\\
\hline
\end{tabular}
\caption{Training matrix characteristics}
\label{Tab1}
\end{center}
\end{table}

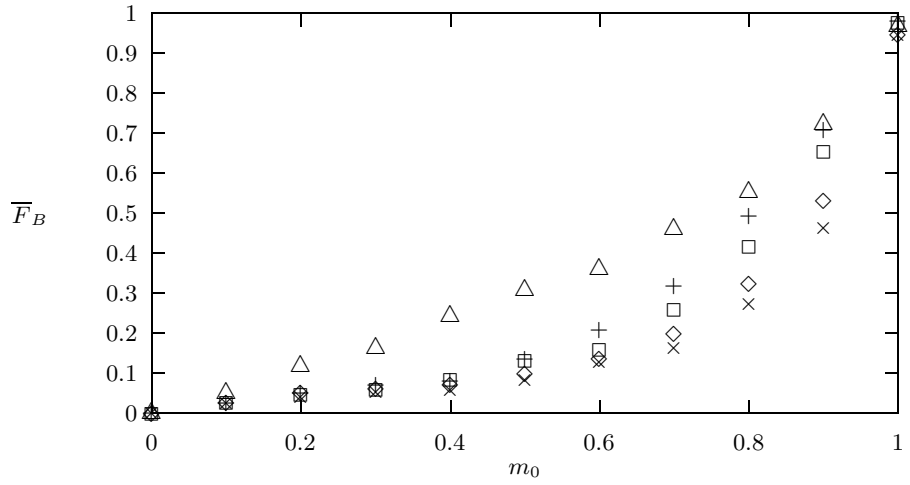
\begin{figure}
\input{FB1.N-10.tex}
\caption{Over generation: Mean fraction of error in recalled patterns $\overline{F}_{B}$ against initial pattern noise $m_{0}$ for 1 bits. T1:$\times$, T2:$\Diamond$, T3:$\Box$, T4:$+$, T5:$\triangle$.}
\label{FB1}
\end{figure}

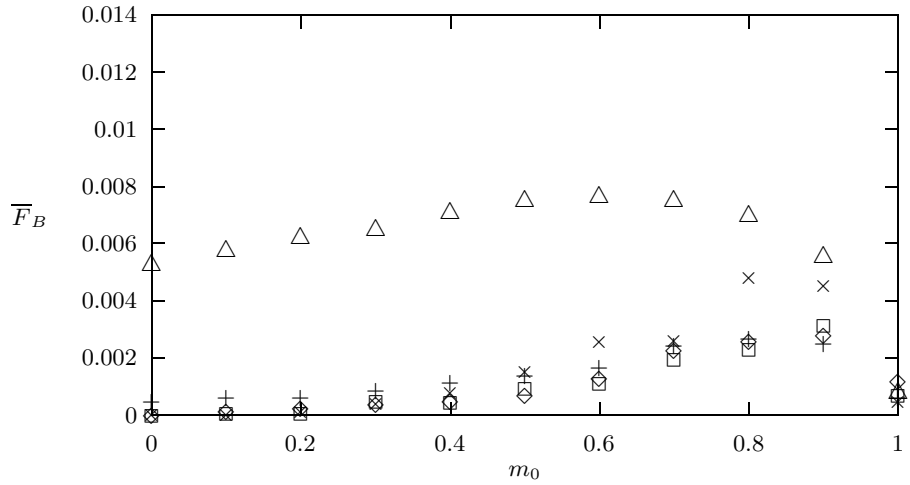
\begin{figure}
\input{FB0.N-10.tex}
\caption{Over generation: Mean fraction of error in recalled patterns $\overline{F}_{B}$ against initial pattern noise $m_{0}$ for 0 bits. T1:$\times$, T2:$\Diamond$, T3:$\Box$, T4:$+$, T5:$\triangle$.}
\label{FB2}
\end{figure}

Due to the greater relevance of recalling a $V^{(\mu)}_{i} = 1$ bit over a $ V^{(\mu)}_{i} = 0$ bit in the nominated image the results are shown in two figures. Figure \ref{FB1} shows $\overline{F}_{B}$ for bits in the nominated image which should be set to 1. Figure \ref{FB2} shows $\overline{F}_{B}$ for $V^{(\mu)}_{i} = 0$ bits.

The mean fraction of randomly flipped bits in the nominated image $V^{(\mu)}$ at the start of processing is shown as $m_{0}$. In all of the simulations except for T5, the mean was calculated from 10 trials of 50 patterns. For T5, 12 trials of 40 patterns were used.

\section{Conclusion}

Since the evaluation approach we use is numerical rather than analytical and we cannot conduct exhaustive tests we must regard the calculations as approximations as far as the storage capacity of the network is concerned. However, we have extrapolated our results over a large number of tests so we are quite confident of at least 2 decimal places of accuracy in the results with an overall error of 0.01 in the mean values for $\overline{F}_{B}$.

Despite the low activation levels shown by sparsity and connectivity in Table \ref{Tab1}, the simulations showed that patterns of natural language sentences could successfully be stored and retrieved from the Hopfield network. No sudden discontinuity in $\overline{F}_{B}$ was observed either for 1 bits or 0 bits.

Overall recall was good for T1 to T4 with a poorer resistance to initial noise by T5. Indeed we note that even in an absence of noise, i.e. when $m_{0}=0.0$, the error in recall of 1 bits for T5 was greater than 0. On close inspection we also see that T4 has a small error at $m_{0}=0.0$. From this evidence we conclude that $\alpha$ for T4 and T5 are above the critical level $\alpha_{c}$ which gives us a value of $0.18 \leq \alpha_{c} < 0.20$ which is in line with Grensing's findings for biased random patterns. This indicates that recall degrades continuously after $\alpha$ exceeds $\alpha_{c}$ upto some point when recall totally fails. In our simulations we have not reached a point of total recall failure.

We should also note that although in general looking at $\overline{F}_{B}$ is not a good way of detecting discontinuities in $\alpha$ we think that in our simulations $N$ is sufficiently large to validate the method.

In line with comments by other researchers (e.g. \cite{amit:89}) we note that one reason for successful recall of patterns in low activity networks is the degree of correlation between patterns. This is despite the values of $\alpha$ exceeding the expected critical value $\alpha_{c}$  Significant correlations between stored patterns could be said to have interacted with bias to increase the critical storage value.

In this paper we have shown that biased patterns which are correlated by an underlying complex linguistic distribution of word cooccurrences can be stored and recalled in a Hopfield network. Moreover, the network behaves differently to one trained with unbiased random patterns in that the critical storage ratio is increased from the theoretical limit and recall degrades continuously. This establishes two basic properties of the Hopfield network for NLP. Once the fundamental behaviour of the Hopfield network is known we can then usefully adapt it to association based NLP.

\newpage
 
\bibliographystyle{plain}

\end{document}

%% file: FB1.N-10.tex
\setlength{\unitlength}{0.240900pt}
\ifx\plotpoint\undefined\newsavebox{\plotpoint}\fi
\sbox{\plotpoint}{\rule[-0.175pt]{0.350pt}{0.350pt}}%
\begin{picture}(1500,900)(0,0)
\sbox{\plotpoint}{\rule[-0.175pt]{0.350pt}{0.350pt}}%
\put(264,158){\rule[-0.175pt]{282.335pt}{0.350pt}}
\put(264,158){\rule[-0.175pt]{0.350pt}{151.526pt}}
\put(264,158){\rule[-0.175pt]{4.818pt}{0.350pt}}
\put(242,158){\makebox(0,0)[r]{0}}
\put(1416,158){\rule[-0.175pt]{4.818pt}{0.350pt}}
\put(264,221){\rule[-0.175pt]{4.818pt}{0.350pt}}
\put(242,221){\makebox(0,0)[r]{0.1}}
\put(1416,221){\rule[-0.175pt]{4.818pt}{0.350pt}}
\put(264,284){\rule[-0.175pt]{4.818pt}{0.350pt}}
\put(242,284){\makebox(0,0)[r]{0.2}}
\put(1416,284){\rule[-0.175pt]{4.818pt}{0.350pt}}
\put(264,347){\rule[-0.175pt]{4.818pt}{0.350pt}}
\put(242,347){\makebox(0,0)[r]{0.3}}
\put(1416,347){\rule[-0.175pt]{4.818pt}{0.350pt}}
\put(264,410){\rule[-0.175pt]{4.818pt}{0.350pt}}
\put(242,410){\makebox(0,0)[r]{0.4}}
\put(1416,410){\rule[-0.175pt]{4.818pt}{0.350pt}}
\put(264,473){\rule[-0.175pt]{4.818pt}{0.350pt}}
\put(242,473){\makebox(0,0)[r]{0.5}}
\put(1416,473){\rule[-0.175pt]{4.818pt}{0.350pt}}
\put(264,535){\rule[-0.175pt]{4.818pt}{0.350pt}}
\put(242,535){\makebox(0,0)[r]{0.6}}
\put(1416,535){\rule[-0.175pt]{4.818pt}{0.350pt}}
\put(264,598){\rule[-0.175pt]{4.818pt}{0.350pt}}
\put(242,598){\makebox(0,0)[r]{0.7}}
\put(1416,598){\rule[-0.175pt]{4.818pt}{0.350pt}}
\put(264,661){\rule[-0.175pt]{4.818pt}{0.350pt}}
\put(242,661){\makebox(0,0)[r]{0.8}}
\put(1416,661){\rule[-0.175pt]{4.818pt}{0.350pt}}
\put(264,724){\rule[-0.175pt]{4.818pt}{0.350pt}}
\put(242,724){\makebox(0,0)[r]{0.9}}
\put(1416,724){\rule[-0.175pt]{4.818pt}{0.350pt}}
\put(264,787){\rule[-0.175pt]{4.818pt}{0.350pt}}
\put(242,787){\makebox(0,0)[r]{1}}
\put(1416,787){\rule[-0.175pt]{4.818pt}{0.350pt}}
\put(264,158){\rule[-0.175pt]{0.350pt}{4.818pt}}
\put(264,113){\makebox(0,0){0}}
\put(264,767){\rule[-0.175pt]{0.350pt}{4.818pt}}
\put(498,158){\rule[-0.175pt]{0.350pt}{4.818pt}}
\put(498,113){\makebox(0,0){0.2}}
\put(498,767){\rule[-0.175pt]{0.350pt}{4.818pt}}
\put(733,158){\rule[-0.175pt]{0.350pt}{4.818pt}}
\put(733,113){\makebox(0,0){0.4}}
\put(733,767){\rule[-0.175pt]{0.350pt}{4.818pt}}
\put(967,158){\rule[-0.175pt]{0.350pt}{4.818pt}}
\put(967,113){\makebox(0,0){0.6}}
\put(967,767){\rule[-0.175pt]{0.350pt}{4.818pt}}
\put(1202,158){\rule[-0.175pt]{0.350pt}{4.818pt}}
\put(1202,113){\makebox(0,0){0.8}}
\put(1202,767){\rule[-0.175pt]{0.350pt}{4.818pt}}
\put(1436,158){\rule[-0.175pt]{0.350pt}{4.818pt}}
\put(1436,113){\makebox(0,0){1}}
\put(1436,767){\rule[-0.175pt]{0.350pt}{4.818pt}}
\put(264,158){\rule[-0.175pt]{282.335pt}{0.350pt}}
\put(1436,158){\rule[-0.175pt]{0.350pt}{151.526pt}}
\put(264,787){\rule[-0.175pt]{282.335pt}{0.350pt}}
\put(45,472){\makebox(0,0)[l]{\shortstack{$\overline{F}_{B}$}}}
\put(850,68){\makebox(0,0){$m_{0}$}}
\put(264,158){\rule[-0.175pt]{0.350pt}{151.526pt}}
\put(264,158){\raisebox{-1.2pt}{\makebox(0,0){$\Diamond$}}}
\put(381,175){\raisebox{-1.2pt}{\makebox(0,0){$\Diamond$}}}
\put(498,191){\raisebox{-1.2pt}{\makebox(0,0){$\Diamond$}}}
\put(616,197){\raisebox{-1.2pt}{\makebox(0,0){$\Diamond$}}}
\put(733,204){\raisebox{-1.2pt}{\makebox(0,0){$\Diamond$}}}
\put(850,221){\raisebox{-1.2pt}{\makebox(0,0){$\Diamond$}}}
\put(967,245){\raisebox{-1.2pt}{\makebox(0,0){$\Diamond$}}}
\put(1084,283){\raisebox{-1.2pt}{\makebox(0,0){$\Diamond$}}}
\put(1202,362){\raisebox{-1.2pt}{\makebox(0,0){$\Diamond$}}}
\put(1319,492){\raisebox{-1.2pt}{\makebox(0,0){$\Diamond$}}}
\put(1436,754){\raisebox{-1.2pt}{\makebox(0,0){$\Diamond$}}}
\put(264,158){\makebox(0,0){$+$}}
\put(381,174){\makebox(0,0){$+$}}
\put(498,190){\makebox(0,0){$+$}}
\put(616,202){\makebox(0,0){$+$}}
\put(733,209){\makebox(0,0){$+$}}
\put(850,244){\makebox(0,0){$+$}}
\put(967,288){\makebox(0,0){$+$}}
\put(1084,358){\makebox(0,0){$+$}}
\put(1202,467){\makebox(0,0){$+$}}
\put(1319,603){\makebox(0,0){$+$}}
\put(1436,774){\makebox(0,0){$+$}}
\sbox{\plotpoint}{\rule[-0.350pt]{0.700pt}{0.700pt}}%
\put(264,158){\raisebox{-1.2pt}{\makebox(0,0){$\Box$}}}
\put(381,175){\raisebox{-1.2pt}{\makebox(0,0){$\Box$}}}
\put(498,187){\raisebox{-1.2pt}{\makebox(0,0){$\Box$}}}
\put(616,196){\raisebox{-1.2pt}{\makebox(0,0){$\Box$}}}
\put(733,212){\raisebox{-1.2pt}{\makebox(0,0){$\Box$}}}
\put(850,241){\raisebox{-1.2pt}{\makebox(0,0){$\Box$}}}
\put(967,258){\raisebox{-1.2pt}{\makebox(0,0){$\Box$}}}
\put(1084,322){\raisebox{-1.2pt}{\makebox(0,0){$\Box$}}}
\put(1202,421){\raisebox{-1.2pt}{\makebox(0,0){$\Box$}}}
\put(1319,570){\raisebox{-1.2pt}{\makebox(0,0){$\Box$}}}
\put(1436,772){\raisebox{-1.2pt}{\makebox(0,0){$\Box$}}}
\sbox{\plotpoint}{\rule[-0.500pt]{1.000pt}{1.000pt}}%
\put(264,159){\makebox(0,0){$\times$}}
\put(381,175){\makebox(0,0){$\times$}}
\put(498,183){\makebox(0,0){$\times$}}
\put(616,191){\makebox(0,0){$\times$}}
\put(733,194){\makebox(0,0){$\times$}}
\put(850,211){\makebox(0,0){$\times$}}
\put(967,239){\makebox(0,0){$\times$}}
\put(1084,260){\makebox(0,0){$\times$}}
\put(1202,329){\makebox(0,0){$\times$}}
\put(1319,449){\makebox(0,0){$\times$}}
\put(1436,752){\makebox(0,0){$\times$}}
\sbox{\plotpoint}{\rule[-0.250pt]{0.500pt}{0.500pt}}%
\put(264,159){\makebox(0,0){$\triangle$}}
\put(381,190){\makebox(0,0){$\triangle$}}
\put(498,233){\makebox(0,0){$\triangle$}}
\put(616,260){\makebox(0,0){$\triangle$}}
\put(733,310){\makebox(0,0){$\triangle$}}
\put(850,352){\makebox(0,0){$\triangle$}}
\put(967,384){\makebox(0,0){$\triangle$}}
\put(1084,448){\makebox(0,0){$\triangle$}}
\put(1202,506){\makebox(0,0){$\triangle$}}
\put(1319,613){\makebox(0,0){$\triangle$}}
\put(1436,767){\makebox(0,0){$\triangle$}}
\end{picture}

%% file: FB0.N-10.tex
\setlength{\unitlength}{0.240900pt}
\ifx\plotpoint\undefined\newsavebox{\plotpoint}\fi
\sbox{\plotpoint}{\rule[-0.175pt]{0.350pt}{0.350pt}}%
\begin{picture}(1500,900)(0,0)
\sbox{\plotpoint}{\rule[-0.175pt]{0.350pt}{0.350pt}}%
\put(264,158){\rule[-0.175pt]{282.335pt}{0.350pt}}
\put(264,158){\rule[-0.175pt]{0.350pt}{151.526pt}}
\put(264,158){\rule[-0.175pt]{4.818pt}{0.350pt}}
\put(242,158){\makebox(0,0)[r]{0}}
\put(1416,158){\rule[-0.175pt]{4.818pt}{0.350pt}}
\put(264,248){\rule[-0.175pt]{4.818pt}{0.350pt}}
\put(242,248){\makebox(0,0)[r]{0.002}}
\put(1416,248){\rule[-0.175pt]{4.818pt}{0.350pt}}
\put(264,338){\rule[-0.175pt]{4.818pt}{0.350pt}}
\put(242,338){\makebox(0,0)[r]{0.004}}
\put(1416,338){\rule[-0.175pt]{4.818pt}{0.350pt}}
\put(264,428){\rule[-0.175pt]{4.818pt}{0.350pt}}
\put(242,428){\makebox(0,0)[r]{0.006}}
\put(1416,428){\rule[-0.175pt]{4.818pt}{0.350pt}}
\put(264,517){\rule[-0.175pt]{4.818pt}{0.350pt}}
\put(242,517){\makebox(0,0)[r]{0.008}}
\put(1416,517){\rule[-0.175pt]{4.818pt}{0.350pt}}
\put(264,607){\rule[-0.175pt]{4.818pt}{0.350pt}}
\put(242,607){\makebox(0,0)[r]{0.01}}
\put(1416,607){\rule[-0.175pt]{4.818pt}{0.350pt}}
\put(264,697){\rule[-0.175pt]{4.818pt}{0.350pt}}
\put(242,697){\makebox(0,0)[r]{0.012}}
\put(1416,697){\rule[-0.175pt]{4.818pt}{0.350pt}}
\put(264,787){\rule[-0.175pt]{4.818pt}{0.350pt}}
\put(242,787){\makebox(0,0)[r]{0.014}}
\put(1416,787){\rule[-0.175pt]{4.818pt}{0.350pt}}
\put(264,158){\rule[-0.175pt]{0.350pt}{4.818pt}}
\put(264,113){\makebox(0,0){0}}
\put(264,767){\rule[-0.175pt]{0.350pt}{4.818pt}}
\put(498,158){\rule[-0.175pt]{0.350pt}{4.818pt}}
\put(498,113){\makebox(0,0){0.2}}
\put(498,767){\rule[-0.175pt]{0.350pt}{4.818pt}}
\put(733,158){\rule[-0.175pt]{0.350pt}{4.818pt}}
\put(733,113){\makebox(0,0){0.4}}
\put(733,767){\rule[-0.175pt]{0.350pt}{4.818pt}}
\put(967,158){\rule[-0.175pt]{0.350pt}{4.818pt}}
\put(967,113){\makebox(0,0){0.6}}
\put(967,767){\rule[-0.175pt]{0.350pt}{4.818pt}}
\put(1202,158){\rule[-0.175pt]{0.350pt}{4.818pt}}
\put(1202,113){\makebox(0,0){0.8}}
\put(1202,767){\rule[-0.175pt]{0.350pt}{4.818pt}}
\put(1436,158){\rule[-0.175pt]{0.350pt}{4.818pt}}
\put(1436,113){\makebox(0,0){1}}
\put(1436,767){\rule[-0.175pt]{0.350pt}{4.818pt}}
\put(264,158){\rule[-0.175pt]{282.335pt}{0.350pt}}
\put(1436,158){\rule[-0.175pt]{0.350pt}{151.526pt}}
\put(264,787){\rule[-0.175pt]{282.335pt}{0.350pt}}
\put(45,472){\makebox(0,0)[l]{\shortstack{$\overline{F}_{B}$}}}
\put(850,68){\makebox(0,0){$m_{0}$}}
\put(264,158){\rule[-0.175pt]{0.350pt}{151.526pt}}
\put(264,158){\raisebox{-1.2pt}{\makebox(0,0){$\Diamond$}}}
\put(381,163){\raisebox{-1.2pt}{\makebox(0,0){$\Diamond$}}}
\put(498,168){\raisebox{-1.2pt}{\makebox(0,0){$\Diamond$}}}
\put(616,174){\raisebox{-1.2pt}{\makebox(0,0){$\Diamond$}}}
\put(733,180){\raisebox{-1.2pt}{\makebox(0,0){$\Diamond$}}}
\put(850,189){\raisebox{-1.2pt}{\makebox(0,0){$\Diamond$}}}
\put(967,216){\raisebox{-1.2pt}{\makebox(0,0){$\Diamond$}}}
\put(1084,259){\raisebox{-1.2pt}{\makebox(0,0){$\Diamond$}}}
\put(1202,273){\raisebox{-1.2pt}{\makebox(0,0){$\Diamond$}}}
\put(1319,283){\raisebox{-1.2pt}{\makebox(0,0){$\Diamond$}}}
\put(1436,211){\raisebox{-1.2pt}{\makebox(0,0){$\Diamond$}}}
\put(264,179){\makebox(0,0){$+$}}
\put(381,184){\makebox(0,0){$+$}}
\put(498,185){\makebox(0,0){$+$}}
\put(616,196){\makebox(0,0){$+$}}
\put(733,209){\makebox(0,0){$+$}}
\put(850,219){\makebox(0,0){$+$}}
\put(967,232){\makebox(0,0){$+$}}
\put(1084,267){\makebox(0,0){$+$}}
\put(1202,278){\makebox(0,0){$+$}}
\put(1319,270){\makebox(0,0){$+$}}
\put(1436,184){\makebox(0,0){$+$}}
\sbox{\plotpoint}{\rule[-0.350pt]{0.700pt}{0.700pt}}%
\put(264,158){\raisebox{-1.2pt}{\makebox(0,0){$\Box$}}}
\put(381,161){\raisebox{-1.2pt}{\makebox(0,0){$\Box$}}}
\put(498,160){\raisebox{-1.2pt}{\makebox(0,0){$\Box$}}}
\put(616,180){\raisebox{-1.2pt}{\makebox(0,0){$\Box$}}}
\put(733,177){\raisebox{-1.2pt}{\makebox(0,0){$\Box$}}}
\put(850,200){\raisebox{-1.2pt}{\makebox(0,0){$\Box$}}}
\put(967,208){\raisebox{-1.2pt}{\makebox(0,0){$\Box$}}}
\put(1084,246){\raisebox{-1.2pt}{\makebox(0,0){$\Box$}}}
\put(1202,261){\raisebox{-1.2pt}{\makebox(0,0){$\Box$}}}
\put(1319,299){\raisebox{-1.2pt}{\makebox(0,0){$\Box$}}}
\put(1436,189){\raisebox{-1.2pt}{\makebox(0,0){$\Box$}}}
\sbox{\plotpoint}{\rule[-0.500pt]{1.000pt}{1.000pt}}%
\put(264,158){\makebox(0,0){$\times$}}
\put(381,158){\makebox(0,0){$\times$}}
\put(498,162){\makebox(0,0){$\times$}}
\put(616,176){\makebox(0,0){$\times$}}
\put(733,192){\makebox(0,0){$\times$}}
\put(850,225){\makebox(0,0){$\times$}}
\put(967,273){\makebox(0,0){$\times$}}
\put(1084,275){\makebox(0,0){$\times$}}
\put(1202,374){\makebox(0,0){$\times$}}
\put(1319,360){\makebox(0,0){$\times$}}
\put(1436,179){\makebox(0,0){$\times$}}
\sbox{\plotpoint}{\rule[-0.250pt]{0.500pt}{0.500pt}}%
\put(264,394){\makebox(0,0){$\triangle$}}
\put(381,416){\makebox(0,0){$\triangle$}}
\put(498,436){\makebox(0,0){$\triangle$}}
\put(616,449){\makebox(0,0){$\triangle$}}
\put(733,475){\makebox(0,0){$\triangle$}}
\put(850,494){\makebox(0,0){$\triangle$}}
\put(967,500){\makebox(0,0){$\triangle$}}
\put(1084,495){\makebox(0,0){$\triangle$}}
\put(1202,470){\makebox(0,0){$\triangle$}}
\put(1319,406){\makebox(0,0){$\triangle$}}
\put(1436,192){\makebox(0,0){$\triangle$}}
\end{picture}

%% file: Nemlap.bbl
\begin{thebibliography}{10}

\bibitem{amit:89}
D.~{Amit}.
\newblock {\em Modeling Brain Function - The world of attractor neural
  networks}.
\newblock Cambridge, England: Cambridge University Press, 1989.

\bibitem{amit:87}
D.~{Amit}, H.~{Gutfeund}, and H.~{Sompolinsky}.
\newblock Information storage in neural networks with low levels of activity.
\newblock {\em Phys. Rev. A}, 35(5):2293+, 1987.

\bibitem{amit:87b}
D.~{Amit}, H.~{Gutfeund}, and H.~{Sompolinsky}.
\newblock Statistical mechanics of neural networks near saturation.
\newblock {\em Ann. Phys.}, 173:30+, 1987.

\bibitem{brown:93}
P.~{Brown}, S.~{Della Pietra}, V.~{Della Pietra}, and R.~{Mercer}.
\newblock The mathematics of statistical machine translation: Parameter
  estimation.
\newblock {\em Computational Linguistics}, 19(2):263--299, 1993.

\bibitem{bruce:87}
A.~{Bruce}, E.~{Gardner}, and D.~{Wallace}.
\newblock Dynamics and statistical mechanics of the {Hopfield} model.
\newblock {\em Journal of Physics A}, 20:2909--2934, 1987.

\bibitem{buhmann:87}
J.~{Buhmann} and K.~{Schulten}.
\newblock Influence of noise on the function of a physiological neural network.
\newblock {\em Biol. Cybern.}, 56:313+, 1987.

\bibitem{collier:95d}
N.~{Collier} and K.~{Takahashi}.
\newblock Sentence alignment in parallel corpora: The {Asahi} corpus of
  newspaper editorials.
\newblock Technical Report 95/11, Centre for Computational Linguistics, UMIST,
  Manchester, UK, October 1995.

\bibitem{forrest:88}
B.~{Forrest}.
\newblock Content-addressability and learning in neural networks.
\newblock {\em Journal of Physics A}, 21:245--255, 1988.

\bibitem{gale:93b}
W.~{Gale}, {Church}. K., and D.~{Yarowsky}.
\newblock A method for disambiguating word senses in a large corpus.
\newblock {\em Computers and the Humanities}, 26:415--439, 1993.

\bibitem{gardner:86}
E.~{Gardner}.
\newblock Structure of metastable states in the {Hopfield} model.
\newblock {\em Journal of Physics A}, 19:L1047--L1052, 1986.

\bibitem{gardner:87}
E.~{Gardner}.
\newblock Multiconnected neural network models.
\newblock {\em Journal of Physics A}, 20:3453--3464, 1987.

\bibitem{grensing:87}
D.~{Grensing}, R.~{K\"{u}hn}, and J.~{van Hammen}.
\newblock Storing patterns in a spin-glass model of neural networks near
  saturation.
\newblock {\em Journal of Physics A}, 20:2935--2947, 1987.

\bibitem{hopfield:82}
J.J. {Hopfield}.
\newblock Neural networks and physical systems with emergent selective
  computational abilities.
\newblock {\em Proceedings of the National Academy of Science, USA}, 79:2554+,
  1982.

\bibitem{hopfield:84}
J.J. {Hopfield}.
\newblock Neurons with graded response have collective computational properties
  like those of two-state neurons.
\newblock {\em Proceedings of the National Academy of Science, USA},
  81:3088--3092, May 1984.

\bibitem{ide:90}
N.~{Ide} and J.~{V\'{e}ronis}.
\newblock Very large neural networks for word sense disambiguation.
\newblock In {\em ECAI90: 9th European Conference on Artificial Intelligence,
  Stockholm, Sweden}, pages 366--368, August 6--10 1990.

\bibitem{kohonen:73}
T.~{Kohonen} and M.~{Rouhounen}.
\newblock Representation of associated data by matrix operators.
\newblock {\em IEEE Trans. Computation}, 22:701, 1973.

\bibitem{nishimori:93}
H.~{Nishimori} and T.~{Ozeki}.
\newblock Retrieval dynamics of associative memory of the {Hopfield} type.
\newblock {\em Journal of Physics A}, 26:859--871, 1993.

\bibitem{ozeki:94}
T.~{Ozeki} and H.~{Nishimori}.
\newblock Noise distributions in retrieval dynamics of the {Hopfield} model.
\newblock {\em Journal of Physics A}, 27:7061--7068, 1994.

\bibitem{personnaz:85}
L.~{Personnaz}, I.~{Guyen}, and G.~{Dreyfus}.
\newblock Information storage and retrieval in spin-glass like neural networks.
\newblock {\em J. Physique Lett.}, 46:L359+, 1985.

\bibitem{tarassenko:90}
L.~{Tarassenko}, B.~{Seifert}, J.~{Tombs}, J.~{Reynolds}, and A.~{Murray}.
\newblock Neural network architectures for associative memory.
\newblock In {\em First IEE International Conference on Artificial Neural
  Networks}, 1990.

\bibitem{watkin:91}
T.~{Watkin} and D.~{Sherrington}.
\newblock The parallel dynamics of a dilute symmetric hebb-rule network.
\newblock {\em Journal of Physics A}, 24:5427--5433, 1991.

\end{thebibliography}
